%% file: omnieval.tex
\documentclass{article}

\usepackage[preprint]{neurips_2025}

\usepackage[utf8]{inputenc} % allow utf-8 input
\usepackage[T1]{fontenc}    % use 8-bit T1 fonts
\usepackage{hyperref}       % hyperlinks
\usepackage{url}% simple URL typesetting
\usepackage{booktabs}       % professional-quality tables
\usepackage{amsfonts}       % blackboard math symbols
\usepackage{nicefrac}       % compact symbols for 1/2, etc.
\usepackage{microtype}      % microtypography
\usepackage{xcolor} % colors
\usepackage{graphicx}
\usepackage{multirow}
\usepackage{booktabs}
\usepackage{todonotes}
\usepackage{amsmath}
\usepackage{floatrow}
\usepackage{subcaption}

\title{OmniEval: A Benchmark for Evaluating Omni-modal Models with Visual, Auditory, and Textual Inputs}

\author{
  Yiman Zhang\textsuperscript{*},
  Ziheng Luo\textsuperscript{*,†},
  Qiangyu Yan\textsuperscript{*}, 
  \\
  \textbf{Wei He\textsuperscript{*},
  Borui Jiang\textsuperscript{*},
  Xinghao Chen\textsuperscript{*,$\S$},
  Kai Han\textsuperscript{*,$\S$}},
  \\
  \textsuperscript{*}Huawei Noah's Ark Lab;
  \textsuperscript{†}University of Science and Technology of China
  \\ % 换到下一行写邮箱
  \textsuperscript{$\S$} Corresponding Author
  \\
  \texttt{\{yiman.zhang, xinghao.chen, kai.han\}@huawei.com}
}

\begin{document}

\maketitle

\begin{abstract}
In this paper, we introduce OmniEval, a benchmark for evaluating omni-modality models like  MiniCPM-O 2.6~\cite{MiniCPMo26GPT4o_misc}, which encompasses visual, auditory, and textual inputs. Compared with existing benchmarks, our OmniEval has several distinctive features: (i) Full-modal collaboration: We design evaluation tasks that highlight the strong coupling between audio and video, requiring models to effectively leverage the collaborative perception of all modalities; (ii) Diversity of videos: OmniEval includes 810 audio-visual synchronized videos, 285 Chinese videos and 525 English videos; (iii) Diversity and granularity of tasks: OmniEval contains 2617 question-answer pairs, comprising 1412 open-ended questions and 1205 multiple-choice questions. These questions are divided into 3 major task types and 12 sub-task types to achieve comprehensive evaluation. Among them, we introduce a more granular video localization task named Grounding. Then we conduct experiments on OmniEval with several omni-modality models. We hope that our OmniEval can provide a platform for evaluating the ability to construct and understand coherence from the context of all modalities. Codes and data could be found at https://omnieval-benchmark.github.io/.
\end{abstract}

\section{Introduction}
The pursuit of Artificial Intelligence (AI) systems capable of emulating human-like understanding of the world has catalyzed significant advancements in models that process information from multiple modalities \cite{radford2021learning,alayrac2022flamingo,li2023blip2}.  These Multimodal Large Language Models (MLLMs) have demonstrated remarkable potential in tasks like image captioning, visual question answering, and text-to-image generation \cite{openaiGPT4TechnicalReport2024, qwenQwen25TechnicalReport2025}. However, a prevailing trend is the development of "omni-modal models" capable of concurrently processing and understanding information from all three modalities: visual, auditory, and textual \cite{xuQwen25OmniTechnicalReport2025,fuVITA15GPT4oLevel2025,chengVideoLLaMA2Advancing2024,MiniCPMo26GPT4o_misc}. Such models aim to more comprehensively simulate human perception and cognition of the world, laying the foundation for more complex and realistic application scenarios, including intelligent assistants, robotic interaction, and content creation.

Despite the promising application prospects of omni-modal models, comprehensively and effectively evaluating their integrated capabilities remains a critical unresolved issue. Existing multimodal benchmarks predominantly focus on combinations of one or two modalities (e.g., vision-text or audio-text) or fail to adequately reflect the deep coupling and synergistic effects among multimodal information in their task design \cite{li2025omnibenchfutureuniversalomnilanguage, hong2025worldsenseevaluatingrealworldomnimodal}.
%x
For instance, some existing benchmarks may focus on static visual content paired with audio, thereby inadequately assessing the understanding of dynamic visual events crucial for real-world scenarios \cite{li2025omnibenchfutureuniversalomnilanguage}. Others, while offering a broader range of tasks, might be limited to a single language, thus failing to evaluate a model's multilingual capabilities \cite{hong2025worldsenseevaluatingrealworldomnimodal}. Consequently, these benchmarks often fall short in evaluating the deep, synergistic understanding that arises from the concurrent integration of dynamic visual, auditory, and textual cues across diverse linguistic contexts. They may also lack the task diversity or the fine-grained evaluation mechanisms, such as precise temporal grounding, necessary to truly probe how omni-modal models interpret and fuse these distinct information streams to achieve a holistic understanding.
Particularly for questions requiring models to simultaneously integrate visual dynamics, sound events, and associated text (such as subtitles or dialogue) for accurate answers, current evaluation methods often prove inadequate. Moreover, existing models still face substantial challenges in real-world understanding, which further underscores the necessity of constructing a more comprehensive and challenging evaluation benchmark.

\begin{figure}[htb]
  \centering
  \includegraphics[width=1.0\textwidth]{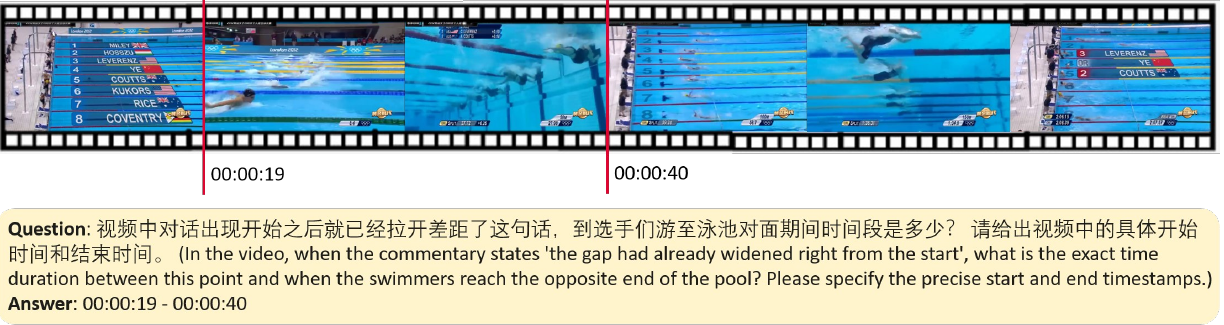}
  \caption{An grounding example in OmniEval. OmniEval requires integrating both visual and auditory signals to provide accurate answers for certain questions, while also incorporating fine-grained understanding tasks such as grounding.}
  \label{fig:example}
\end{figure}

To address this critical evaluation gap, we introduce OmniEval, a novel benchmark specifically designed to rigorously evaluate omni-modal models that jointly process and reason across visual, auditory, and textual inputs, supporting both Chinese and English languages. 
OmniEval possesses several distinctive features: 1) \textbf{Full-modal Collaborative Evaluation:} We have meticulously designed evaluation tasks that emphasize the strong coupling between audio and video, requiring models to effectively leverage the collaborative perception of all modalities for correct answers (Figure~\ref{fig:example}). This transcends evaluation approaches that merely sum individual unimodal understanding capabilities. 2) \textbf{Diverse Videos and Task Scenarios:} OmniEval comprises 810 audio-visual synchronized video clips, including 285 Chinese videos and 525 English videos. These videos ensuring broad coverage of evaluation scenarios. 3) \textbf{Diverse and Fine-grained Task Design:} OmniEval contains 2617 question-answer pairs, consisting of 1412 open-ended questions and 1205 multiple-choice questions. These questions are divided into 3 major task types and 12 sub-task types, aiming for a comprehensive assessment of model capabilities. Notably, we introduce a more fine-grained video localization task, termed Grounding (Figure~\ref{fig:example}), to precisely evaluate the model's ability to locate information in the temporal dimension.

Based on OmniEval, we have conducted extensive evaluations of various state-of-the-art omni-modal models. The experimental results indicate that existing models face significant challenges in understanding real-world information. This clearly demonstrates the challenging nature of OmniEval and the urgent need to enhance the capabilities of current models.

The main contributions of this paper are as follows:
\begin{itemize}
    \item We construct and release OmniEval, a novel and comprehensive omni-modal evaluation benchmark suite, that focuses on assessing models' synergistic understanding and processing of visual, auditory, and textual information, with bilingual support (Chinese and English).
    \item OmniEval introduces diverse video content and fine-grained task types, particularly establishing tasks that emphasize strong audio-visual coupling and precise temporal localization (Grounding), offering a new perspective for a more comprehensive measurement of model capabilities. %\todo{Check the description of task features and grounding in contribution 2.}
    \item We conduct extensive testing and analysis of current mainstream omni-modal models on OmniEval, providing valuable baselines, revealing the deficiencies of existing models in real-world understanding, and offering insights for future research directions. %\todo{check details of experients and benchmark}
\end{itemize}

We hope that OmniEval will serve as an important benchmark to drive the development of omni-modal models, encouraging researchers to build more powerful models capable of understanding and constructing coherence from the context of all modalities. Our dataset and evaluation code are publicly available to foster further research in the community.

\section{Related Work}
\label{Related}

\subsection{Multimodal Large Language Models}
%大语言模型取得了较好的结果
%当前已有一些工作试图将更多的模态融合在大语言模型中。
%在视觉方面，这些模型通过一个精心设计的编码器对视觉信息进行编码，并通过Q-former或MLP在语义层面与文本token进行对齐。之后利用大语言模型强大的生成能力，在多个任务上取得了良好的效果。
%在听觉方面（介绍qwen audio）
%除此之外，一些更新的工作试图使用一个模型处理文本、视觉与听觉
%然而，如何评估这些模型的能力仍然是一个亟待解决的问题。
Recent advancements in large language models (LLMs) have demonstrated significant improvements across a wide range of natural language processing (NLP) tasks\cite{deepseek-aiDeepSeekV3TechnicalReport2025,openaiGPT4TechnicalReport2024,brownLanguageModelsAre2020,qwenQwen25TechnicalReport2025}. These models, characterized by their deep architectures and extensive pretraining on massive corpora, have consistently outperformed traditional methods in benchmarks such as question answering\cite{BONFIGLI2024103003}, machine translation\cite{zhang2024survey}, summarization\cite{BONFIGLI2024103003}, and text generation\cite{openaiGPT4TechnicalReport2024,brownLanguageModelsAre2020}.
There has been an increasing interest in incorporating multiple modalities into large language models (LLMs), with the goal of enhancing their capabilities beyond textual processing alone.\cite{li2023blip2,liuVisualInstructionTuning2023,xuQwen25OmniTechnicalReport2025,fuVITA15GPT4oLevel2025}
In the visual domain, raw images are processed through specialized visual encoders to obtain high-level features, while in the audio domain, raw waveforms are first sampled and then encoded using dedicated audio encoders. These modality-specific representations are subsequently aligned with textual tokens using intermediate modules such as Querying Transformers (Q-Former)\cite{li2023blip2}, Multi-Layer Perceptrons (MLPs)\cite{liuVisualInstructionTuning2023}, or other alignment techniques\cite{wangCogVLMVisualExpert2024}. This semantic alignment enables the fusion of heterogeneous inputs into a unified representation space. Leveraging the generative capabilities of LLMs, the resulting multimodal architecture achieves strong performance across a range of tasks, including image captioning\cite{liuVisualInstructionTuning2023,wangCogVLMVisualExpert2024}, visual and spoken question answering, audio captioning\cite{chuQwenAudioAdvancingUniversal2023}, and multimodal dialogue\cite{fuVITA15GPT4oLevel2025}.
In addition, some models have attempted to integrate both visual and auditory understanding into a single, unified framework, thereby creating omni-modality models\cite{xuQwen25OmniTechnicalReport2025,fuVITA15GPT4oLevel2025,chengVideoLLaMA2Advancing2024,MiniCPMo26GPT4o_misc}. However, evaluating the performance of such models presents a significant challenge, as it requires designing tasks that simultaneously involve multiple modalities. The lack of standardized evaluation metrics and benchmarks for these models remains an open problem, and addressing this issue is critical for advancing the development and comparison of multimodal AI systems.

\subsection{Multimodal Benchmarks}
% 当前已有很多benchmark来评估大语言模型的理解、推理等能力。在视觉方面，之前的工作在物体识别与定位、对图像提问的理解与回答、视觉常识与常识推理等多个维度对模型能力进行评估。在听觉方面，已有工作在自动语音识别、语音指令跟随、语音问答、音频场景理解等多个维度对模型能力进行评估。
%针对大语言模型，当前已有很多benchmark，用来评测模型的推理、等能力。
%对于视觉大模型，当前已有很多benchmark，用来评测模型的定位、推理、ocr、等能力。
%在语音方面，当前已有很多benchmark，用来评测模型的asr、等能力
%对于同时涉及音频与视频的，当前benchmark较少同时还存在一定缺陷
%omnibench提供语音和图片，测试模型对语音内容的理解能力和静态视觉的感知能力，缺少对动态视觉内容理解能力的测试。
%worldsense提出了一个相对全面的benchmark，测试模型的count、等能力。然而，worldsense只针对单个语言，忽略了其他语言。
%针对以上情况，我们提出了OmniEval，一个全面的全模态模型benchmark，针对多个场景、多个任务、多种语言对模型开展更为全面的评测。

Recently, a wide range of benchmarks exist to evaluate the understanding and reasoning capabilities of large language models\cite{zellersHellaSwagCanMachine2019,wangMMLUProMoreRobust2024,hendrycksMeasuringMassiveMultitask2021,cobbeTrainingVerifiersSolve2021}. In the visual domain, prior works assess model performance across multiple dimensions, including object recognition\cite{flickr30k,flickrentitiesijcv,Li-hallucination-2023} and localization\cite{kazemzadehReferItGameReferringObjects2014,yuModelingContextReferring2016}, image-based question answering\cite{balanced_vqa_v2,AntolVQA2015,balanced_binary_vqa,liuMMBenchYourMultimodal2024a,gurariVizWizGrandChallenge2018}, and visual commonsense reasoning\cite{masry-etal-2022-chartqa,luMathVistaEvaluatingMathematical2024,singhVQAModelsThat2019}. Similarly, in the auditory domain, existing benchmarks focus on tasks such as automatic speech recognition\cite{Hernandez_2018,conneau2022fleursfewshotlearningevaluation,VassilLibrispeech2015,bu2017aishell1opensourcemandarinspeech,zhang2022wenetspeech10000hoursmultidomain,Chen_2021}, audio-based question answering\cite{joshi2017triviaqalargescaledistantly,lipping2022clothoaqacrowdsourceddatasetaudio,nachmani2024spoken,yang2024airbenchbenchmarkinglargeaudiolanguage}, and audio scene understanding\cite{poriaMELDMultimodalMultiParty2019,chenEmotionLinesEmotionCorpus2018,Nagrani_2017,yang2024airbenchbenchmarkinglargeaudiolanguage}. These benchmarks serve as essential tools for measuring the effectiveness of multimodal models in real-world applications, enabling systematic comparisons across different modalities and model architectures.
For omni-modality models, the number of available benchmarks is limited, and most of them exhibit certain shortcomings. Some studies\cite{li2025omnibenchfutureuniversalomnilanguage} provide static images along with speech to test models' abilities in speech understanding and static scene perception, yet they overlook the model's capacity to process dynamic visual information. Other work\cite{hong2025worldsenseevaluatingrealworldomnimodal} focuses on testing models' understanding of both audio and video by providing video and audio inputs, but these benchmarks often lack diversity in testing scenarios, tasks, and languages. As a result, there is a need for more comprehensive and standardized evaluation frameworks that can better assess the full range of capabilities in omni-modality models, including their ability to handle dynamic multimodal inputs across varied real-world conditions.
To address these issues, we propose OmniEval, a comprehensive benchmark designed specifically for evaluating the full range of capabilities in omni-modality models.

\input{sections/3_method.tex}

\begin{table}[htbp]
\centering
\caption{\textbf{Overall performance on OmniEval. MNT indicates max new tokens. OE indicates open-ended QAs, MC indicates multiple-choice QAs.}}
\label{main results on OmniEval}
\resizebox{\linewidth}{!}{
\begin{tabular}{@{}l|cccccccccccc@{}}
\toprule
\multirow{2}{*}{\textbf{Methods}}                     & \multirow{2}{*}{\textbf{Params}} & \multirow{2}{*}{\textbf{Frames}} & \multirow{2}{*}{\textbf{MNT}} & \multicolumn{2}{c}{\textbf{Perception}} & \multicolumn{2}{c}{\textbf{Understanding}} & \multicolumn{2}{c}{\textbf{Reasoning}} & \multicolumn{2}{c}{\textbf{Avg}} & \multirow{2}{*}{\textbf{Overall}} \\ \cmidrule(r){5-6} \cmidrule(r){7-8} \cmidrule(r){9-10} \cmidrule(r){11-12}
                                                      &                                  &                                  &                               & OE                 & MC                 & OE                   & MC                  & OE                 & MC                & OE              & MC             &                                   \\ 
\hline
Qwen2.5-Omni-7B\cite{xuQwen25OmniTechnicalReport2025}      & 7B                      & 1fps                    & 1024                 & 53.93          & 67.50         & 49.45           & 65.60           & 65.57         & 88.90         & 53.64      & 69.68      & 61.02                       \\
Qwen2.5-Omni-3B\cite{xuQwen25OmniTechnicalReport2025}                      & 3B                      & 1fps                    & 1024                 & 50.93          & 65.00         & 42.35           & 59.10           & 60.60         & 88.30         & 48.73      & 65.61      & 56.50                       \\
Baichuan-Omni-1.5\cite{li2025baichuan}         & 7B                      & 64                      & 1024                 & 39.12          & 61.90         & 34.63           & 61.70           & 48.47         & 85.40         & 38.53      & 65.12      & 50.77                      \\
MiniCPM-O 2.6\cite{MiniCPMo26GPT4o_misc}               & 8B                      & 64                      & 1024                 & 28.70          & 26.67         & 27.08           & 31.59           & 20.22         & 25.73         & 26.95      & 28.77      & 27.79                     \\
VITA-1.5\cite{fuVITA15GPT4oLevel2025}                 & 8B                      & 64                      & 1024                 & 5.27           & 11.88         & 10.28           & 7.22            & 4.26          & 8.77          & 7.15       & 9.29       & 8.14                        \\
\hline
gemini-2.5-pro-preview-05-06~\cite{gemini2.5}                & -                       & 1fps                    & -                    & 54.04          & 71.00         & 63.38           & 68.40           & 81.42         & 60.20         & 61.34      & 68.27      & 64.56      \\                
\bottomrule
\end{tabular}}
\end{table}

\begin{table}[tbp]
\centering
\caption{\textbf{Performance of the model on different language dimensions on OmniEval.}}
\label{encn results on OmniEval}
\resizebox{\linewidth}{!}{
\begin{tabular}{@{}l|cccccccccc}
\toprule
\multirow{2}{*}{\textbf{Methods}}    & \multirow{2}{*}{\textbf{Params}} & \multirow{2}{*}{\textbf{Frames}} & \multirow{2}{*}{\textbf{MNT}} & \multicolumn{3}{c}{\textbf{English}} & \multicolumn{3}{c}{\textbf{Chinese}} \\
\cmidrule(r){5-7} \cmidrule(r){8-10} 
                              &                         &                         &                      & OE      & MC      & ALL     & OE      & MC      & ALL     \\
\hline
Qwen2.5-Omni-7B\cite{xuQwen25OmniTechnicalReport2025}     & 7B                      & 1fps                    & 1024                 & 48.89   & 70.88   & 60.17   & 58.81   & 67.52   & 62.19   \\
Qwen2.5-Omni-3B\cite{xuQwen25OmniTechnicalReport2025}    & 3B                      & 1fps                    & 1024                 & 45.92   & 65.98   & 56.21   & 51.80   & 64.95   & 56.90   \\
Baichuan-Omni-1.5\cite{li2025baichuan}         & 7B                      & 64                      & 1024                 & 41.59   & 64.43   & 53.30   & 35.19   & 66.36   & 47.28   \\
MiniCPM-O 2.6\cite{MiniCPMo26GPT4o_misc}       & 8B                      & 64                      & 1024                 & 12.12   & 13.31   & 12.73   & 43.09   & 58.34   & 49.00   \\
VITA-1.5\cite{fuVITA15GPT4oLevel2025}       & 8B                      & 64                      & 1024                 & 2.35    & 0.40    & 1.35    & 12.38   & 25.51   & 17.47   \\
\hline
gemini-2.5-pro-preview-05-06~\cite{gemini2.5}        & -                       & 1fps                    & -                    & 61.71   & 68.56   & 65.22   & 60.95   & 67.76   & 63.59     \\
\bottomrule
\end{tabular}}
\end{table}

\section{Experiments and Findings}
\label{Exper}
% 本节将基于 OmniEval 基准对当前开源的 MLLM 进行全面评估。我们首先阐述了实验方法和评估方案，然后对定量结果进行了全面的分析。此外，我们还对影响性能的重要因素进行了详细的研究，从而为多模态理解的潜在方向提供了一些见解。
In this section, we conduct a comprehensive evaluation of existing open-source multimodal MLLMs and Gemini 2.5\cite{gemini2.5} based on the proposed OmniEval benchmark. We begin by outlining the experimental setup and evaluation methodology, detailing the tasks, metrics and data used in our analysis. We then present an in-depth examination of the quantitative results, highlighting the strengths and weaknesses of different models across various modalities and tasks. Furthermore, we investigate several key factors that influence model performance, offering insights into the challenges and opportunities in multimodal understanding. 
%Our findings aim to inform future research directions and support the development of more robust and generalizable omni-modality models.
\subsection{Settings}

% 为了全面评估多模态理解能力，我们评估了x种类型的 MLLM：

To comprehensively evaluate the multimodal understanding capabilities of current models, we assess 6 fully multimodal models that integrate visual, textual, and auditory information. The evaluation configuration parameters are shown in Table \ref{main results on OmniEval}.

For evaluation, we adopt different strategies for MC and OE Q\&A pairs. For MC Q\&A pairs, we directly determine whether the option output by the model is consistent with the ground truth. For OE questions, we leverage a powerful proprietary language model to assist in assessment. Specifically, we categorize Q\&A pairs into grounding, counting and other tasks and utilize different assessment methods for different categories.

% the inverse of the video’s frame rate (1/FPS) or the ratio of video duration to max\_frame. 

\subsubsection{Evaluation for Grounding OE Q\&A pairs}
%For Grounding open-ended tasks, LLM is firstly used to extract temporal information from the model output. Then different strategies will be used to evaluate different types of data.  Specifically, for moment-based Q\&A pairs, we construct an adaptive evaluation method based on the video frame extraction strategy. When the number of extracted frames is very low, the time intervals between adjacent frames become significantly larger. In such cases, the model's prediction and the true value may not align precisely. Therefore, a larger threshold should be employed to evaluate the model's output, allowing for a more lenient assessment of correctness. As described in Eq.\ref{ts val}, the answer is deemed correct if the difference falls within a predefined threshold. This threshold is determined by either FPS(Frames Per Second) or the max frame number and the video duration. 
For grounding open-ended tasks, we first leverage large language models (LLMs) to extract temporal information from the model's output. Subsequently, we employ distinct strategies to evaluate various data types.

Specifically, for moment-based Q\&A pairs, we've developed an adaptive evaluation method based on video frame extraction. When the number of extracted frames is low, the time intervals between adjacent frames become significantly larger. In such scenarios, precise alignment between the model's prediction and the true value may not be achievable. Therefore, we use a larger threshold to evaluate the model's output, allowing for a more lenient assessment of correctness. As shown in Eq.\ref{ts val}, an answer is considered correct if the difference falls within this predefined threshold, which is determined by either the frames per second (FPS) or a combination of the maximum frame number and video duration.

\begin{equation}
\label{ts val}
\text{R} =
\begin{cases}
    \text{True}, & \text{if } |\hat{t} - t_{\text{gt}}| \leq \tau_{ts} \\
    \text{False}, & \text{otherwise} 
\end{cases}
\text{, where }
\tau_{ts} = \min\left(\frac{1}{\text{FPS}}, \frac{\text{video\_duration}}{\text{max\_frame}}\right)
\end{equation}

where $R$ indicated the discriminant result, $\hat{t}$ indicates the time stamp extracted from the model output, $t_{\text{gt}}$ indicates the ground truth time stamp and $\tau_{ts}$ indicates the threshold.

%As described in Eq.\ref{timespan val}, similar to LongVALE~\cite{geng2024longvale}, for time span-based open-ended Grounding Q\&A pairs, we use the Intersection over Union(IoU) between the predicted and ground-truth time intervals to determine whether the answer is correct or not.
Similar to LongVALE~\cite{geng2024longvale}, for time span-based open-ended Grounding Q\&A pairs, we evaluate correctness using the Intersection over Union (IoU) between the predicted and ground-truth time intervals, as detailed in Equation \ref{timespan val}.
For our evaluation, $ \tau_{time\_span}$ was set to 0.5.
\begin{equation}
\label{timespan val}
\text{R} =
\begin{cases}
    \text{True}, & \text{if } \text{IoU}(\hat{I}, I_{\text{gt}}) \geq \tau_{time\_span} \\
    \text{False}, & \text{otherwise}
\end{cases}
\end{equation}

\iffalse
\begin{equation}
\text{IoU}(\hat{I}, I_{\text{gt}}) = 
\frac{|\hat{I} \cap I_{\text{gt}}|}{|\hat{I} \cup I_{\text{gt}}|}
\end{equation}
\fi

where  $\hat{I}$ indicates the time span extracted from the model output, $I_{\text{gt}}$ indicates the ground truth time span and $\tau_{time\_span}$ indicates the threshold.

\subsubsection{Evaluation for Counting and other OE Q\&A pairs}
% For counting open-ended tasks, such as Object Counting and Action Counting, large language models (LLMs) are employed to precisely extract numerical values from the model outputs. Subsequently, these extracted values are then compared with the ground truth. If they match, the response is deemed correct; otherwise, it is considered incorrect.

% For other open-ended tasks, we utilize large models to calculate the similarity between the model outputs and the ground truth, and assign scores to the answers based on this similarity. The scores are floating-point numbers between 0 and 1, where 1 indicates a completely correct answer and 0 signifies a completely incorrect one.
For counting open-ended tasks, like object or action counting, LLMs are used to precisely extract numerical values from the model outputs. These extracted values are then directly compared to the ground truth: a match indicates a correct response, while any mismatch is considered incorrect.

For other open-ended tasks, we leverage LLMs to compute the similarity between the model outputs and the ground truth. Answers are then assigned a score, a floating-point number between 0 and 1, where 1 signifies a completely correct answer and 0 denotes a completely incorrect one.

\subsection{Results on OmniEval}

\textbf{Main Results.} 
%We present the overall evaluation results on OmniEval in Table \ref{main results on OmniEval} and Table \ref{encn results on OmniEval}, covering the performance of various MLLMs across three target categories—perception, understanding, and reasoning—two question formats—MC and OE—and two languages—English and Chinese.
The comprehensive evaluation results on OmniEval are presented in Table \ref{main results on OmniEval} and \ref{encn results on OmniEval}. Table \ref{main results on OmniEval} details MLLM performance across three target categories (perception, comprehension, and reasoning), whereas Table \ref{encn results on OmniEval} highlights language-specific performance (English and Chinese). Both tables further delineate MLLM performance on open-ended (OE) and multiple-choice (MC) question formats.
% 下面介绍一些结果上的发现

% As Table \ref{main results on OmniEval} shows, gemini-2.5-pro-preview-05-06 achieved the highest overall score of 64.40, leading particularly in Reasoning OE (80.87) and Perception MC (71.00). Qwen2.5-Omni 7B followed with an overall score of 61.19, generally outperforming its 3B counterpart (56.66). Baichuan-Omni 1.5, another 7B model, scored 51.06, showing stronger MC performance than OE. MiniCPM-O 2.6 (37.87 overall) and VITA-1.5 (11.61 overall) showed comparatively lower performance. Worth noting, VITA-1.5 encounters tensor size out-of-bounds issues when simultaneously receiving video and audio information for samples exceeding approximately 200 seconds in length.

As Table \ref{main results on OmniEval} shows, gemini-2.5-pro-preview-05-06 achieved the highest overall score of 64.56, leading particularly in Perception MC (71.00) and Reasoning OE (81.42). Qwen2.5-Omni-7B followed with an overall score of 61.02, generally outperforming its 3B counterpart (specifically, Qwen2.5-Omni-3B with 1fps achieved 56.50 overall, and Baichuan-Omni 1.5 with 64 frames achieved 50.77 overall). MiniCPM-O 2.6 scored 27.79 overall, and ViTA-1.5 scored 8.14 overall, showing comparatively lower performance. It is worth noting that ViTA-1.5 encounters tensor size out of range issues when receiving video and audio information with a sample length of over about 200 seconds simultaneously. In addition, Minicpm-o also encounters size mismatch issues on some test cases.

% Gemini-2.5-pro-preview-05-06 also excels in both English and Chinese, with overall scores of 65.22 (EN) and 63.59 (CN), showcasing robust bilingual capabilities. MiniCPM-O 2.6 demonstrates a unique strength in Chinese, achieving a higher Chinese overall score (56.96) than English (12.73), largely driven by its Chinese OE performance (46.48). Other models like Qwen2.5-Omni and Baichuan-Omni 1.5 show varying degrees of proficiency in both languages.
Gemini-2.5-pro-preview-05-06 demonstrates robust bilingual capabilities, achieving 65.22 (EN) and 63.59 (CN) overall scores. MiniCPM-O 2.6 uniquely excels in Chinese (49.00 overall, driven by 58.34 MC) compared to English (12.73). Qwen2.5-Omni models perform strongly in both languages (7B: 60.17 EN, 62.19 CN; 3B: 56.21 EN, 56.90 CN). Baichuan-Omni-1.5 shows moderate performance (53.30 EN, 47.28 CN), while ViTA-1.5 lags significantly (1.35 EN, 17.47 CN).

These results underscore the advanced capabilities of models like Gemini 2.5 Pro on complex multi-modal tasks, highlighting their superior performance and robust bilingual support on OmniEval.

\subsection{Impact of Visual Information and Audio Information}
% 鉴于先前评估中观察到的显著性能差异，我们进一步研究了不同类型的模态特定数据如何影响多模态大型语言模型 (MLLM) 的整体性能。
In light of the significant performance disparities observed in the preceding evaluation, we further investigate how different types of modality-specific data contribute to the overall performance of  open-source MLLMs. Specifically, we analyze the impact of visual, auditory, and multimodal inputs on task outcomes, aiming to understand the relative importance and interplay of each modality. This exploration provides valuable insights into the data composition and modality balance required for effective multimodal understanding.

\subsubsection{Visual Information.}
% To analyze the contribution of visual information, we experiment with three input settings: audio-only, audio with captions, and audio with visual frames. 
% Table \ref{Impact of visual Information} consistently indicates that incorporating captions significantly enhances model performance across all evaluated methods. For instance, Qwen2.5-Omni 7B's overall score increased from 48.39 (Audio-only) to 61.95 (Audio with captions). Conversely, the subsequent addition of raw video frames generally does not yield further improvements, and in some cases, leads to performance degradation. This observation, particularly evident in models like MiniCPM-O 2.6 (Overall: 60.86 to 37.87), suggests that textual captions are more effectively leveraged than raw video content by these MLLMs in the current evaluation setup.

To assess the contribution of visual information, experiments were conducted across three input modalities: audio-only, audio augmented with captions, and audio augmented with visual frames.  As presented in Table \ref{Impact of visual Information}, the incorporation of captions consistently yields a substantial enhancement in model performance across all evaluated methods. For example, Qwen2.5-Omni-7B exhibited an increase in overall score from 47.91 (audio-only) to 61.67 (audio with captions). Conversely, the subsequent addition of raw video frames generally did not lead to further improvements; in several instances, it resulted in performance degradation. This phenomenon, notably observed with MiniCPM-O 2.6 (overall score decreasing from 56.54 to 27.79 with video addition), suggests that under the current evaluation paradigm, these MLLMs more effectively leverage textual captions than raw video content.

\begin{table}[htbp]
\vspace{-1.0em}
\centering
\caption{Impact of visual information for MLLMs.}
\label{Impact of visual Information}
\resizebox{\linewidth}{!}{
\begin{tabular}{l|cccccccccccc}
\toprule
\multirow{2}{*}{\textbf{Methods}} & \multicolumn{3}{c}{\textbf{Perception}} & \multicolumn{3}{c}{\textbf{Understanding}} & \multicolumn{3}{c}{\textbf{Reasoning}} & \multicolumn{3}{c}{\textbf{Overall}} \\ \cmidrule(r){2-4} \cmidrule(r){5-7} \cmidrule(r){8-10} \cmidrule(r){11-13} 
     & Audio       & +Caption     & +Video     & Audio& +Caption      & +Video      & Audio      & +Caption     & +Video     & Audio      & +Caption    & +Video    \\ 
\hline
Qwen2.5-Omni-7B\cite{xuQwen25OmniTechnicalReport2025}   & 46.62   & 65.81      & 60.18   & 40.55    & 52.96       & 56.89    & 75.67   & 75.83     & 76.31   & 47.91  & 61.67     & 61.02  \\
Qwen2.5-Omni-3B\cite{xuQwen25OmniTechnicalReport2025}  & 47.22   & 67.98      & 57.41   & 40.03    & 50.45       & 50.06    & 72.18   & 75.29     & 73.36   & 47.47  & 61.49     & 56.50  \\
Baichuan-Omni-1.5\cite{li2025baichuan}      & 44.50   & 51.71      & 49.61   & 38.04    & 42.05       & 47.09    & 68.59   & 53.28     & 65.47   & 45.11  & 47.93     & 50.77  \\
MiniCPM-O 2.6\cite{MiniCPMo26GPT4o_misc}    & 46.17   & 64.81      & 27.76   & 36.82    & 47.45       & 29.15    & 59.81   & 59.02     & 22.76   & 43.96  & 56.54     & 27.79  \\
VITA-1.5\cite{fuVITA15GPT4oLevel2025}       & 19.85   & 27.93      & 8.31    & 14.97    & 24.66       & 8.87     & 22.40   & 24.62     & 6.34    & 18.05  & 26.07     & 8.14   \\   

\bottomrule
\end{tabular}}
\end{table}

\subsubsection{Audio Information.}
%We investigate the impact of audio information through different input configurations: (1) video frames only, (2) video frames with video subtitles, and (3) video frames with audio. 
%The results in Table \ref{Impact of Audio Information} consistently demonstrate that incorporating subtitles with video significantly enhances model performance across all methods and categories. For instance, Qwen2.5-Omni 7B's overall score increased notably from 58.07 (Video-only) to 64.59 (Video + Subtitle). Conversely, the addition of raw audio to video yields mixed results: while some models show improvements (e.g., Baichuan-Omni 1.5 overall from 43.24 to 51.06), others exhibit limited gains or even performance degradation (e.g., MiniCPM-O 2.6's overall score decreased from 43.10 to 37.87 when audio was added). This suggests that the multimodal understanding of audio by these multimodal language models (MLLMs) still has room for improvement.

To assess audio information's impact, we evaluated three input configurations: video-only, video+subtitles, and video+audio.  Table \ref{Impact of Audio Information}  demonstrates that subtitles consistently enhance performance (e.g., Qwen2.5-Omni-7B's overall score increased from 49.96 to 64.40). Conversely, adding raw audio yields mixed results; some models improve (e.g., Baichuan-Omni-1.5 overall: 43.06 to 50.77), while others degrade (e.g., MiniCPM-O 2.6 overall: 54.88 to 27.79). This indicates that the multimodal understanding of raw audio by current MLLMs still requires significant advancement.

\begin{table}[htbp]
\vspace{-1.0em}
\centering
\caption{Impact of audio information for MLLMs.}
\label{Impact of Audio Information}
\resizebox{\linewidth}{!}{
\begin{tabular}{l|cccccccccccc}
\toprule
\multirow{2}{*}{\textbf{Methods}} & \multicolumn{3}{c}{\textbf{Perception}} & \multicolumn{3}{c}{\textbf{Understanding}} & \multicolumn{3}{c}{\textbf{Reasoning}} & \multicolumn{3}{c}{\textbf{Overall}} \\ \cmidrule(r){2-4} \cmidrule(r){5-7} \cmidrule(r){8-10} \cmidrule(r){11-13} 
     & Video      & +Subtitle     & +Audio     & Video       & +Subtitle      & +Audio      & Video      & +Subtitle     & +Audio    & Video     & +Subtitle    & +Audio    \\ \midrule
Qwen2.5-Omni-7B\cite{xuQwen25OmniTechnicalReport2025}& 51.89   & 63.07      & 60.18   & 47.42    & 60.08       & 56.89    & 50.63  & 81.50      & 76.31   & 49.96  & 64.40     & 61.02  \\
Qwen2.5-Omni-3B\cite{xuQwen25OmniTechnicalReport2025} & 45.82   & 60.72      & 57.41   & 42.06    & 57.65       & 50.06    & 45.20  & 78.84      & 73.36   & 44.21  & 61.98     & 56.50  \\
Baichuan-Omni-1.5\cite{li2025baichuan}            & 42.35   & 49.18      & 49.61   & 42.49    & 46.23       & 47.09    & 46.09  & 65.63      & 65.47   & 43.06  & 50.20     & 50.77  \\
MiniCPM-O 2.6\cite{MiniCPMo26GPT4o_misc}             & 44.46   & 55.89      & 27.76   & 38.40    & 50.28       & 29.15    & 38.59  & 66.23      & 22.76   & 41.13  & 54.88     & 27.79  \\
VITA-1.5\cite{fuVITA15GPT4oLevel2025}              & 10.58   & 11.24      & 8.31    & 10.39    & 12.47       & 8.87     & 6.52   & 6.60       & 6.34    & 9.80   & 10.97     & 8.14  \\ \bottomrule
\end{tabular}}
\end{table}
\vspace{-15pt}

\section{Conclusion}

%In this paper, we introduce OmniEval, a refined video understanding benchmark designed to address limitations of current evaluation datasets. OmniEval offers bilingual support (English and Chinese), enabling direct assessment of multilingual omni-modal models, a key departure from predominantly English-centric benchmarks. Its significant inclusion of open-ended questions provides a more thorough evaluation of generative capabilities compared to MCQ-heavy benchmarks. Furthermore, the explicit and granular integration of event grounding allows for targeted assessment of a model's ability to link answers to specific video moments, surpassing coarser temporal methods. This benchmark serves as a valuable and complementary resource for the research community, fostering more nuanced progress in video understanding.
In this paper, we introduced OmniEval, a refined video understanding benchmark meticulously designed to address the significant limitations of current evaluation methodologies. OmniEval distinguishes itself through several key contributions: its inherent bilingual support (English and Chinese) enables the crucial direct evaluation of multilingual omni-modal models, a capability largely absent in predominantly English-centric benchmarks. Furthermore, the benchmark's substantial inclusion of open-ended questions facilitates a more comprehensive and nuanced assessment of Omni-modal Large Models' generative capabilities, moving beyond the constraints of benchmarks heavily reliant on multiple-choice formats. Finally, the explicit and granular integration of event grounding provides a targeted evaluation of these models' ability to precisely connect answers to specific video moments, thereby advancing beyond coarser temporal localization approaches. Collectively, OmniEval offers a valuable and complementary resource for the research community, fostering more nuanced and holistic progress in the challenging domain of video understanding.

\newpage
\bibliographystyle{plain}
\bibliography{omnieval}

\end{document}

%% file: sections/3_method.tex
\section{OmniEval}
\label{omnieval_construct}

\begin{table}[tbp]
\caption{The comparison of various benchmarks encompasses several key aspects: Modality involved(\textbf{Modality}), Languages involved(\textbf{Language}),  Format of Q\&A pair(\textbf{QA Format}), whether including Event Grounding Task(\textbf{Grounding}), the source of videos(\textbf{video Sources}), the method of generating questions and answers(\textbf{QA Generation}) and the number of Q\&A pairs(\textbf{No. of QA Pairs}). A, V, I for modality represent audio, video, and image. OE indicates open-ended questions, MC indicates multiple-choice questions.}
\label{Comparison_of_benchmark}
\resizebox{\linewidth}{!}{
\begin{tabular}{l|ccccccc}
\hline
\textbf{Feature}                            & \textbf{Modality} & \textbf{Language} & \textbf{QA Format} & \textbf{Grounding}       & \textbf{Video Sources} & \textbf{QA Generation} & \textbf{No. of QA Pairs} \\ \hline
OmniBench~\cite{li2024omnibench}            & I+A               & EN                & MC                 & No                       & No                     & Manual                 & 1143                     \\
MMbench-Video~\cite{fang2024mmbench}        & V                 & EN                & OE                 & No                       & YouTube                & Manual                 & 1998                     \\
DeVE-QA~\cite{qin2024question}              & V                 & EN                & Limited OE         & Yes (Grounding required) & ActivityNet            & LLM + Manual         & 78000                    \\ \hline
Video-MME~\cite{fu2024video}                & V+A               & EN                & MC                 & No                       & YouTube                & Manual                 & 2700                     \\

WorldSense~\cite{hong2025worldsense}        & V+A               & EN                & MC                 & Yes(Coarse-grained)      & YouTube, MusicAVQA     & Manual                 & 3172                     \\
LongVALE~\cite{geng2024longvale}            & V+A               & EN                & No QA              & Yes                      & YouTube                & LLM + Manual           & 0                        \\
StreamingBench~\cite{lin2024streamingbench} & V+A               & EN                & OE                 & No                       & YouTube                & LLM + Manual           & 4500                     \\
CG-Bench~\cite{chen2024cg}                  & V+A               & EN                & MC                 & No                       & YouTube, BiliBili      & Manual Curation        & 12129                    \\
\hline
\textbf{OmniEval}                           & V+A               & EN \& CN          & MC \& OE           & Yes (Fine-grained)       & YouTube,Youku,Bilibili & LLM + Manual           & 2617                     \\ \hline
\end{tabular}}
\end{table}
%\vspace{-20pt}

\subsection{Motivation and Overview}
% Recent advancements in Multimodal Large Language Models (Omni models) have significantly improved video understanding capabilities.1 However, existing benchmarks often exhibit limitations, primarily focusing on English-language content and frequently relying on multiple-choice question (MCQ) formats, which may not fully capture the generative and nuanced reasoning abilities of advanced models. For instance, prominent benchmarks like WorldSense, LongVALE, StreamingBench, and VideoMM predominantly utilize English videos and QA pairs. Furthermore, while MCQs facilitate standardized evaluation 1, they may not adequately assess a model's capacity for open-ended generation and detailed explanation.4
To address these gaps and foster more comprehensive MLLM evaluation, we introduce a new benchmark dataset specifically designed with multilingual support and a balanced mix of question formats. This chapter details the systematic pipeline developed for its construction, emphasizing methodological rigor and quality control. Our pipeline integrates automated data processing using large models with essential manual curation, aiming to create a challenging and reliable resource for evaluating Omni models across diverse cognitive tasks, including fine-grained event understanding inspired by the need for temporal localization.

\subsection{Data Collection and Preprocessing Pipeline}
This phase focused on assembling a diverse video collection and extracting the necessary textual modalities (captions and speech transcripts) to serve as the foundation for Q\&A generation.

We initiated the process by aggregating video information from a variety of sources, including established video benchmarks (FineVideo~\cite{Farre2024FineVideo}, Youku-mplug~\cite{xu2023youku}) and diverse web platforms such as bilibili. This hybrid sourcing strategy aimed to ensure broad coverage of topics, styles, and real-world scenarios, moving beyond the confines of specific dataset domains. The goal was to create a varied collection challenging models on multiple fronts.

For each identified video, we acquired corresponding captions with Qwen2.5-VL-70B. When available from source benchmarks, existing high-quality caption tracks were utilized. For other videos, captions were obtained or generated using appropriate methods, ensuring a textual description accompanied each video.

To capture the linguistic content within the audio track, we employed the Volcano Engine large model for Automated Speech Recognition (ASR). Accurate ASR was performed for all videos, configured for the primary languages present (Chinese and English), yielding transcripts of spoken content.

A crucial filtering step was applied based on the ASR transcripts. Videos containing little or no spoken content (identified via metrics like word count or speech duration) were excluded. Specifically, as with FineVideo\cite{Farre2024FineVideo}, we calculate the word sensitivity for each video and exclude those with subdensity less than 0.5. This ensures that the remaining videos possess meaningful linguistic information in the audio modality, complementing the visual stream. This step is vital as our subsequent Q\&A generation leverages both captions and transcripts (Section 3.3), aiming to probe deeper audio-visual understanding rather than purely visual recognition.

\subsection{Q\&A Pair Generation and Annotation Pipeline}
Leveraging the curated videos and their associated text, we implemented a multi-stage pipeline for generating and categorizing QA pairs.

\begin{figure}[htbp]
    \centering
    \begin{subfigure}[b]{0.6\textwidth}
        \includegraphics[width=\textwidth]{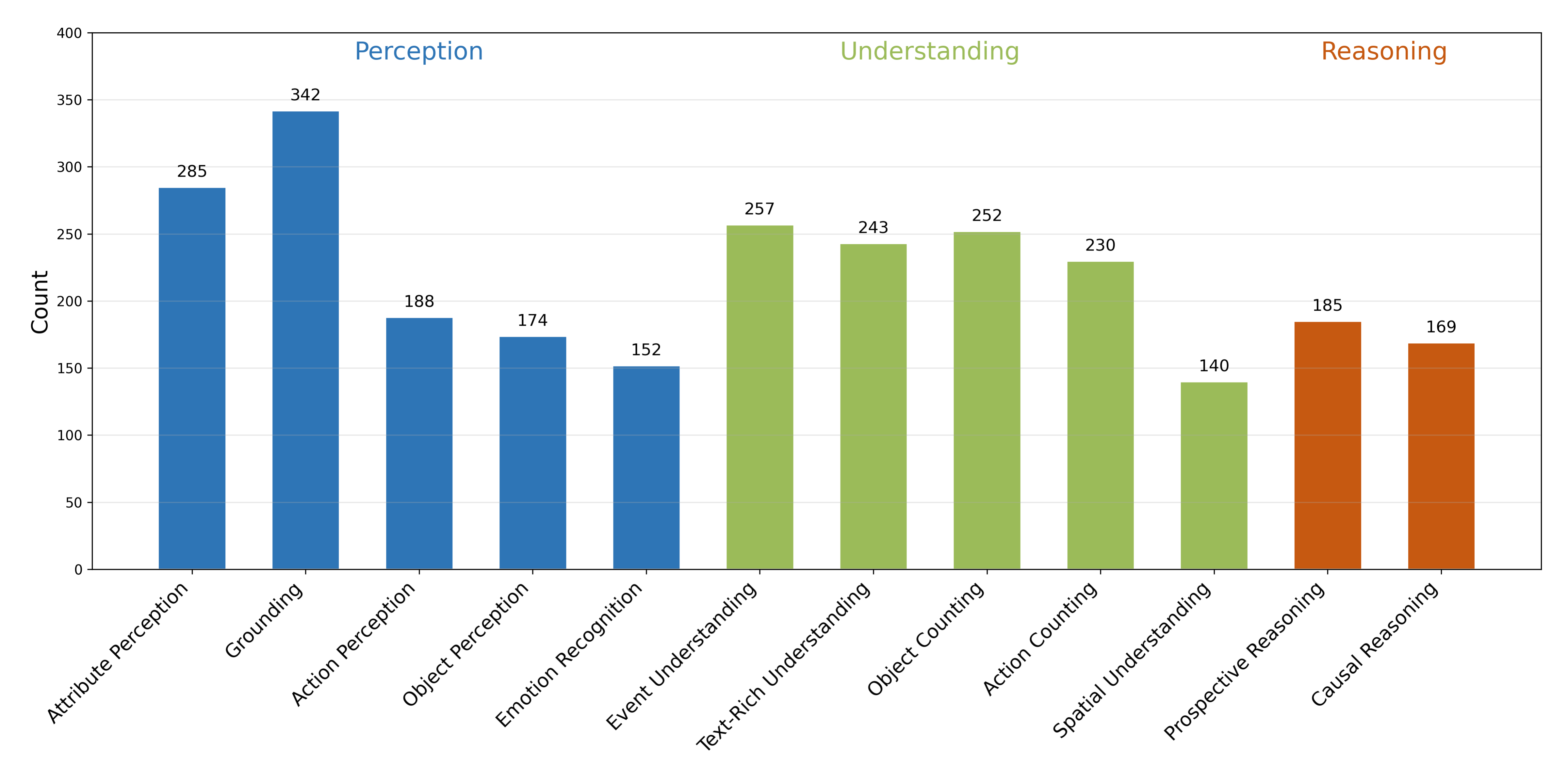}
        %\caption{Task question quantity grouped by functional categories in OmniEval.    Blue bars represent perception-related tasks, green indicates information processing tasks,   and orange denotes higher-order reasoning tasks.}
        \label{fig:tasks}
    \end{subfigure}
    \hfill % 添加一些水平间距
    \begin{subfigure}[b]{0.35\textwidth}
        \includegraphics[width=\textwidth]{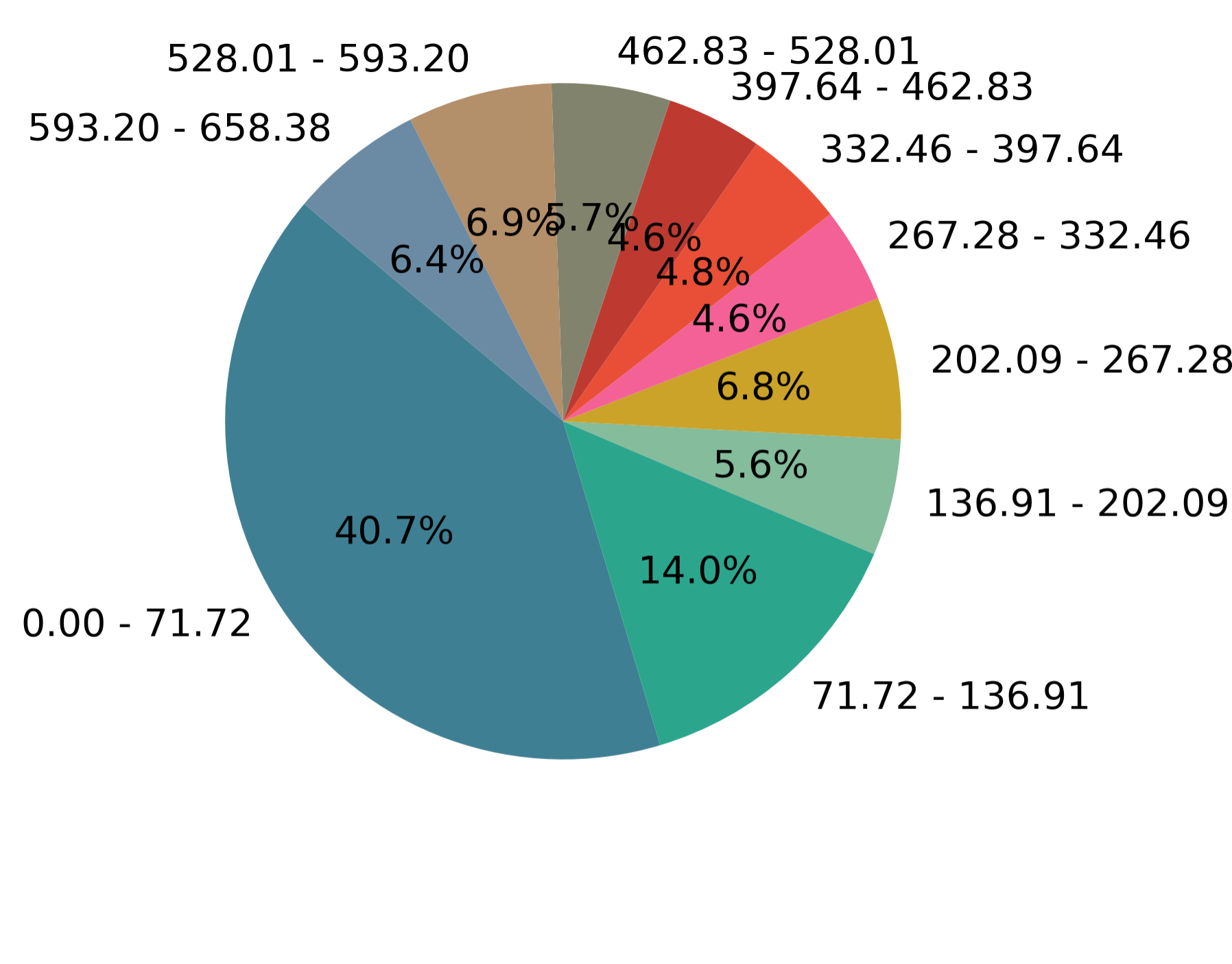}
        %\caption{Video duration distribution. OmniEval covers videos of various lengths. The average video length in the dataset is 211 seconds.}
        \label{fig:video_dur}
    \end{subfigure}
    \caption{The left diagram depicts task question quantity grouped by functional categories in OmniEval. Blue bars represent perception-related tasks, green indicates information processing tasks,   and orange denotes higher-order reasoning tasks. The right diagram depicts video duration distribution. The unit of the number is second. OmniEval covers videos of various lengths. The average video length in the dataset is 211 seconds.}
    \label{fig:main}
\end{figure}

\floatsetup[table]{capposition=top}
\newfloatcommand{capbtabbox}{table}[][\FBwidth]
\begin{table*}
\begin{floatrow}
\capbtabbox{
\begin{tabular}{lc}
\hline
\textbf{Question Format}      & \textbf{Num.} \\
\hline
Open-Ended(OE)      & 1412  \\
Multiple-Choice(MC) & 1205  \\
Total                & 2617  \\
\hline
\end{tabular}
}{
 \caption{Format Distribution of Q\&A pairs in OmniEval.}
 \label{Tab1}
}
\capbtabbox{
\begin{tabular}{lccl}
\hline
\textbf{Language} & \textbf{Videos Num.} & \textbf{Q\&A Pairs Num.} \\
\hline
Chinese (CN)      & 285                      & 1104                        \\
English (EN)      & 525                       & 1513                        \\
Total             & 810                       & 2617                        \\
\hline
\end{tabular}
}{
 \caption{Language distribution of videos and Q\&A pairs in OmniEval.}
 \label{Tab2}
 \small
}
\end{floatrow}
\end{table*}

\subsubsection{Automated Q\&A Generation}
We employed large models (LLMs/MLLMs) for automated Q\&A generation, capitalizing on their ability to process multimodal context and formulate relevant questions.1 The process involved three stages:
\begin{itemize}
\item Open-Ended (OE) Generation (Step 4): Models were prompted with both video captions and audio subtitles to generate OE questions and corresponding answers. This approach provides rich context, combining descriptive text with spoken dialogue/narration. Generating OE questions first allows for capturing more complex and nuanced aspects of the video content without the initial constraint of predefined answer choices.
\item Multiple-Choice (MC) Derivation (Step 5): Subsequently, the generated OE pairs were used as input for another large model task: converting the OE question into an MC format. This involved generating plausible distractors alongside the correct answer derived from the OE pair. Including MC questions facilitates standardized evaluation protocols common in the field.
\item Removing those overly simple samples (Step 6): To ensure the complexity and robustness of the benchmark, we rigorously evaluated the Q\&A pairs using multiple large models, and systematically removed questions that could be answered correctly by all models. This process helps to maintain a high level of challenge within the benchmark.
\end{itemize}

Specifically, we have meticulously crafted two distinct categories of Q\&A pairs tailored for \textbf{Grounding}: moment-based and time span-based. Moment questions zero in on pinpointing the precise instant when fleeting events unfold within the video, exemplified by queries like, “At what exact moment does the girl in red commence her speech within the frame?” Conversely, time span questions delve into the broader temporal context, seeking to identify the specific duration during which a particular event transpires, such as, “Over which interval in the video does the girl in red engage in delivering her speech?”. Each grounding Q\&A pair is categorized into either moment-based or the time span-based category and is assessed using different methods accordingly.

\subsubsection{Q\&A Classification}
Each generated Q\&A pair (both OE and MC) was automatically classified into one of 12 predefined categories reflecting different cognitive skills: Grounding, Object Counting, Action Counting, Prospective Reasoning, Text-Rich Understanding, Event Understanding, Attribute Perception, Action Perception, Spatial Understanding, Causal Reasoning, Object Perception, and Emotion Recognition. This fine-grained classification enables nuanced analysis of model strengths and weaknesses across different facets of multimodal understanding, The inclusion of a grounding category specifically targets the model's ability to link answers to specific temporal moments in the video.

\subsubsection{Manual Curation and Quality Assurance}
Recognizing the potential limitations of fully automated generation 4, the final and most critical stage involved meticulous manual review and revision of all Q\&A pairs by human annotators. This step was essential to guarantee:

\begin{itemize}
\item Clarity: Refining question and answer wording for unambiguity. 
\item Relevance and Grounding: Confirming questions are pertinent and answerable from the video, not just based on model biases.
\item Accuracy: Ensuring answers are factually correct based on video content.
\item Judgement: To determine the number of modalities of information required to answer a question correctly and refine the task type of questions.
\item Distribution: Given that the Q\&A pair directly generated by large language models are unevenly distributed in terms of capability items, such as Grounding, Action Counting, Object Counting, we asked five people to watch the videos and write corresponding question-answer pairs.
\end{itemize}

This human-in-the-loop verification significantly enhances the benchmark's reliability and validity, ensuring it genuinely tests multimodal comprehension derived from the video source material.

\subsubsection{Benchmark Statistics}
The construction pipeline yielded a benchmark with a significant number of Q\&A pairs distributed across different formats, task types, and languages.

%As Table~\ref{Tab1} and Table~\ref{Tab2} shows, this benchmark offers a well-balanced distribution between the open-ended (OE) and multiple-choice (MC) question formats, thereby effectively accommodating a variety of evaluation criteria. When a large language model is available, it enables the analysis of performance on both OE and MC questions. Conversely, in the absence of a large language model to assist in evaluating answers to OE questions, the benchmark still allows for a thorough analysis of performance on MC questions alone.

As shown in Tables \ref{Tab1} and \ref{Tab2}, our benchmark features a well-balanced distribution of open-ended (OE) and multiple-choice (MC) question formats, accommodating diverse evaluation criteria. This design enables performance analysis on both OE and MC questions when an LLM is available for OE evaluation. Conversely, without an LLM for OE assistance, the benchmark still facilitates a thorough analysis of MC question performance alone.

The question–answer pairs are classified into 12 distinct types, enabling a fine-grained analysis of model performance across various cognitive skills. Special attention is given to the inclusion of a grounding task (342 pairs), which addresses the need for models to precisely localize information in the temporal dimension.

\iffalse
\begin{figure}[htb]
  \centering
  \includegraphics[width=0.9\textwidth]{fig/tasks1.png}
  \caption{Task question quantity grouped by functional categories in OmniEval.   Blue bars represent perception-related tasks, green indicates information processing tasks,   and orange denotes higher-order reasoning tasks.}
  \label{fig:tasks}
\end{figure}
\fi

A key characteristic of our benchmark is its bilingual nature, encompassing both Chinese (CN) and English (EN) videos and Q\&A pairs. This facilitates research in multilingual MLLM capabilities.

\subsection{Comparison with Existing Benchmarks}
Our benchmark introduces several distinguishing features compared to existing video understanding benchmarks, aiming to provide a more comprehensive evaluation tool for omni models. Table~\ref{Comparison_of_benchmark} provides a comparative overview.

Key differentiators of our benchmark include:

\begin{itemize}
\item Bilingual Support: Unlike many prominent benchmarks that are predominantly English-based (e.g., WorldSense, LongVALE, StreamingBench), our benchmark incorporates a significant volume of both English and Chinese videos and Q\&A pairs. This facilitates direct evaluation and development of omni models for these two major languages.
\item Emphasis on Open-Ended Questions: Many existing benchmarks heavily rely on MCQs for evaluation (e.g. WorldSense, DeVE-QA). Our benchmark provides a substantial number of OE questions (1412 pairs), allowing for a more in-depth assessment of omni models' generative capabilities, their ability to formulate detailed explanations, and their performance in scenarios that mimic natural human interaction more closely than restricted choice formats. 
\item Integrated Event Grounding: While benchmarks like LongVALE and DeVE-QA emphasize temporal understanding and event localization, our benchmark uniquely includes grounding as one of its 12 Q\&A categories. This enables targeted evaluation of a model's ability to connect answers to specific video segments, demonstrating comprehension beyond mere pattern matching. Although WorldSense features coarse-grained "Temporal Localization" multiple-choice questions (e.g., event at beginning/middle/end), our grounding questions offer both multiple-choice and open-ended formats, targeting exact video moments with greater granularity and an adaptive evaluation strategy.
%While benchmarks like LongVALE and DeVE-QA  emphasize temporal understanding and event localization, our benchmark explicitly includes grounding as one of the 12 Q\&A categories. This allows for a targeted evaluation of a model's ability to connect its answers to specific segments or moments within the video, which is crucial for demonstrating true comprehension beyond pattern matching. It is worth noting that WorldSense also features multiple-choice questions categorized under "Temporal Localization." However, these questions are relatively coarse-grained, merely determining whether an event occurs at the beginning, middle, end of the video, or throughout the entire process. In contrast, our grounding questions encompass both multiple-choice and open-ended formats, targeting the exact moment of the video with greater granularity and provide an adaptive strategy to evaluate models.
\end{itemize}

By addressing these aspects, our benchmark aims to complement existing resources and provide a more nuanced and comprehensive platform for advancing MLLM research in video understanding.